\documentclass[sigconf]{acmart}

\setcopyright{none}              % remove ACM copyright block
\renewcommand\footnotetextcopyrightpermission[1]{} % remove footnote
\acmDOI{}                        % clear DOI
\acmISBN{}                       % clear ISBN

\usepackage{tabularx} % for X-column auto-width
\newcolumntype{L}{>{\raggedright\arraybackslash}X} % define ragged column type

\AtBeginDocument{%
  }

\copyrightyear{2025}
\acmYear{2025}
\setcopyright{licensedusgovmixed}\acmConference[CIKM '25]{Proceedings of the 34th ACM International Conference on Information and Knowledge Management}{November 10--14, 2025}{Seoul, Republic of Korea}
\acmBooktitle{Proceedings of the 34th ACM International Conference on Information and Knowledge Management (CIKM '25), November 10--14, 2025, Seoul, Republic of Korea}
\acmDOI{10.1145/3746252.3761487}
\acmISBN{979-8-4007-2040-6/2025/11}

\begin{document}

%%
%% The "title" command has an optional parameter,
%% allowing the author to define a "short title" to be used in page headers.
\title{An Interventional Approach to Real-Time Disaster Assessment via Causal Attribution}

%%
%% The "author" command and its associated commands are used to define
%% the authors and their affiliations.
%% Of note is the shared affiliation of the first two authors, and the
%% "authornote" and "authornotemark" commands
%% used to denote shared contribution to the research.

% \author{
%     Saketh Vishnubhatla\textsuperscript{1},
%     Alimohammad Beigi\textsuperscript{1},
%     Rui Heng Foo\textsuperscript{1},
%     Umang Goel\textsuperscript{1},
%     Ujun Jeong\textsuperscript{1},
%     Bohan Jiang\textsuperscript{1},
%     Adrienne Raglin\textsuperscript{2},
%     Huan Liu\textsuperscript{1} \\
%     \textsuperscript{1}School of Computing and Augmented Intelligence, Arizona State University \\
%     \textt{\{svishnu6, abeigi, rfoo1, ugoel1, ujeong1, bjiang14, huanliu\}@asu.edu} \\
%     \textsuperscript{2}DEVCOM Army Research Lab, USA \\
%     \textt{adrienne.raglin2.civ@army.mil}
% }

\author{Saketh Vishnubhatla}
\orcid{0009-0005-9288-1777}
\affiliation{%
  \institution{School of Computing and Augmented Intelligence, Arizona State University}
  \city{Tempe}
  \state{AZ}
  \country{USA}
}
\email{svishnu6@asu.edu}

\author{Alimohammad Beigi}
\orcid{0009-0009-6637-0761}
\affiliation{%
  \institution{School of Computing and Augmented Intelligence, Arizona State University}
  \city{Tempe}
  \state{AZ}
  \country{USA}
}
\email{abeigi@asu.edu}

\author{Rui Heng Foo}
\orcid{0009-0000-0560-0302}
\affiliation{%
  \institution{School of Computing and Augmented Intelligence, Arizona State University}
  \city{Tempe}
  \state{AZ}
  \country{USA}
}
\email{rfoo1@asu.edu}

\author{Umang Goel}
\orcid{0009-0002-6048-7642}
\affiliation{%
  \institution{School of Computing and Augmented Intelligence, Arizona State University}
  \city{Tempe}
  \state{AZ}
  \country{USA}
}
\email{ugoel1@asu.edu}

\author{Ujun Jeong}
\orcid{0000-0001-5467-0374}
\affiliation{%
  \institution{School of Computing and Augmented Intelligence, Arizona State University}
  \city{Tempe}
  \state{AZ}
  \country{USA}
}
\email{ujeong1@asu.edu}

\author{Bohan Jiang}
\orcid{0000-0001-8552-2681}
\affiliation{%
  \institution{School of Computing and Augmented Intelligence, Arizona State University}
  \city{Tempe}
  \state{AZ}
  \country{USA}
}
\email{bjiang14@asu.edu}

\author{Adrienne Raglin}
\orcid{0000-0002-5489-1633}
\affiliation{%
  \institution{DEVCOM Army Research Lab}
  \city{Adelphi}
  \state{MD}
  \country{USA}
}
\email{adrienne.raglin2.civ@army.mil}

\author{Huan Liu}
\orcid{0000-0002-3264-7904}
\affiliation{%
  \institution{School of Computing and Augmented Intelligence, Arizona State University}
  \city{Tempe}
  \state{AZ}
  \country{USA}
}
\email{huanliu@asu.edu}

%%
%% By default, the full list of authors will be used in the page
%% headers. Often, this list is too long, and will overlap
%% other information printed in the page headers. This command allows
%% the author to define a more concise list
%% of authors' names for this purpose.
\renewcommand{\shortauthors}{Vishnubhatla et al.}

%%
%% The abstract is a short summary of the work to be presented in the
%% article.
\begin{abstract}
Traditional disaster analysis and modelling tools for assessing the severity of a disaster are predictive in nature. Based on the past observational data, these tools prescribe how the current input state (e.g., environmental conditions, situation reports) results in a severity assessment. However, these systems are not meant to be interventional in the causal sense, where the user can modify the current input state to simulate counterfactual ``what-if'' scenarios. In this work, we provide an alternative interventional tool that complements traditional disaster modelling tools by leveraging real-time data sources like satellite imagery, news, and social media. Our tool also helps understand the causal attribution of different factors on the estimated severity, over any given region of interest. In addition, we provide actionable recourses that would enable easier mitigation planning. Our source code\footnote{\url{https://github.com/sak-18/disaster-assessment-tool-UI.git}} is publicly available.%

\end{abstract}

%%
%% The code below is generated by the tool at http://dl.acm.org/ccs.cfm.
%% Please copy and paste the code instead of the example below.

% \begin{CCSXML}
% <ccs2012>
% <concept>
% <concept_id>10010147.10010341.10010349.10010360</concept_id>
% <concept_desc>Computing methodologies~Interactive simulation</concept_desc>
% <concept_significance>500</concept_significance>
% </concept>
% <concept>
% <concept_id>10010147.10010178.10010187.10010192</concept_id>
% <concept_desc>Computing methodologies~Causal reasoning and diagnostics</concept_desc>
% <concept_significance>500</concept_significance>
% </concept>
% </ccs2012>
% \end{CCSXML}



%%
%% Keywords. The author(s) should pick words that accurately describe
%% the work being presented. Separate the keyw rds with commas.
%% A "teaser" image appears between the author and affiliation
%% information and the body of the document, and typically spans the
%% page.
% \begin{teaserfigure}
%   \includegraphics[width=\textwidth]{sampleteaser}
%   \caption{Seattle Mariners at Spring Training, 2010.}
%   \Description{Enjoying the baseball game from the third-base
%   seats. Ichiro Suzuki pre aring to bat.}
%   \label{fig:teaser}
% \end{teaserfigure}

% \received{20 February 2007}
% \received[revised]{12 March 2009}
% \received[accepted]{5 June 2009}

%%
%% This command processes the author and affiliation and title
%% information and builds the first part of the formatted document.

\maketitle

\section{Introduction}

One of the important aspects of disaster management is disaster assessment, which includes estimating the losses to assets and human lives. Assessing the current impact of a disaster helps devise a suitable response strategy. Contemporary disaster modeling tools that help disaster response teams include tools like HAZUS \cite{fema2020hazus} and NASA's disaster mapping tool \cite{nasaDisasters2023}. HAZUS estimates losses in terms of asset damages, economic losses, along with the cost-effectiveness of various mitigation strategies, and is helpful for pre-event analysis. However, it is not intended for real-time assessment. NASA's disaster mapping tool maps different damages on the map, providing a before-and-after view of the disaster for real-time monitoring. However, many of these real-time disaster monitors do not quantify the contribution of various factors (e.g., infrastructural damage, tree cover loss) to the event's severity or provide users with interventional recommendations for mitigating the impact.  

\begin{figure}[!htbp]
    \centering    \includegraphics[width=0.9\columnwidth]{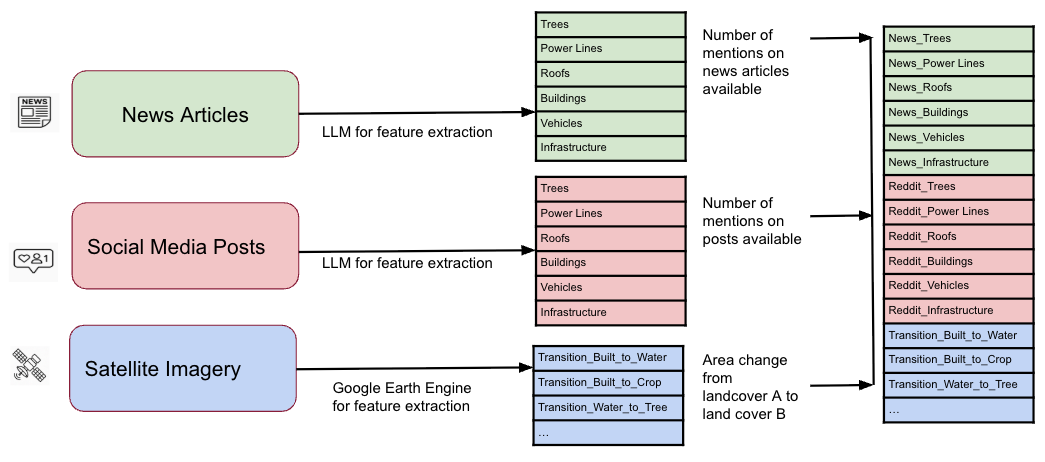} % replace with your filename
    \caption{Illustration of the data curation pipeline.}
    \label{fig:architecture}
\end{figure}

Recently, a growing body of literature has emerged on using remote sensing as a data source for disaster assessment. Other useful data sources include news reports, which offer global event-level information, while social media provides real-time updates on the ground during a disaster. Using these data sources, we plan to develop an interventional system for disaster assessment that allows users to modify current input parameters and simulate the impact on the assessment. In our work, we propose: (1) an interventional system that simulates hypothetical scenarios as $do(.)$ interventions on different causal graphs pertaining to disaster severity assessment; (2) causal attribution score implementation that calculates the necessity and sufficiency of different factors at a feature, feature group, and data source level; and (3) a recourse implementation that provides users with recommendations to change the current severity assessment to a desired state.

\section{Data Collection and Modelling}
For our proposed system, we use three different sources of data: remote sensing, news reports, and social media posts related to billion-dollar disasters for the calendar year 2017. For a given event, our units of analysis are counties within the U.S. We use Dynamic World \cite{brown2022dynamic} land cover maps for remote sensing to compute area changes between different land covers during the disaster. Further, we use a collection of articles from Google News relevant to the disasters and social media posts from Reddit. Our dataset curation mostly follows from our previous work \cite{vishnubhatla2025impact}.

%\vspace{-2 mm}

\subsection{Dataset Preparation}

Our data contains records for 867 unique counties, recording 9 different billion-dollar events and totaling 1097 unique event-county combinations. Below, we outline the features extracted for the counties of interest from various sources.\\
\textbf{Remote Sensing}: For each county, we extract different features corresponding to land cover transitions from a source class to a target class (e.g., area of built-up land in the county that changed to water cover). The feature set includes 81 features, capturing all pairwise transitions between each of the nine land cover classes in Dynamic World. Each feature indicates the area change over the course of the event in meter-squared. \\
\textbf{News}: We use our previously curated news articles from the Google News feed based on keywords specific to the disaster of interest and the county name. Using GPT-4o, we extract key features from each article indicating mentions of damages to powerlines, roofs, infrastructure damages, trees, roads, and bridges. We aggregate the mentions from all the articles (e.g., total powerline damage mentions) in the county to build six features. \\ 
\textbf{Reddit}: We gather relevant posts from the pushshift Reddit archive as a data source. ``r/LocationReddits'' maps subreddits to their locations. Similar to news article curation, we extract key features from each relevant post indicating mentions of damages to powerlines, roofs, infrastructure damages, trees, roads, and bridges. To extract features, we aggregate the mentions from all posts in a county.
%\vspace{-2 mm}

\subsection{Causal DAGs}
To account for varying relationships between different features, we provide three different Causal DAGs. Firstly, we categorise our features into different feature groups that align logically. Then, to build the DAG, we connect all features from one source group to the features in the target group. We consider three different causal graph structures for disaster severity assessment: DAG 1 (Independent Effects), where each feature group independently affects the target (severity); DAG 2 (Mediation through Infrastructure), where flood surface increase and vegetation losses affect infrastructure, which in turn impacts mobility, this pattern is common in disasters like hurricanes or wildfires, where storm surges and fallen trees block roads and damage buildings; and DAG 3 (Flood as Root Cause), where flood damage is treated as the root cause influencing other feature groups, which subsequently affect the severity.

\subsection{Model Evaluation}
We train different models on severity assessment for property damage. As ground truth, we rely on data from SHELDUS (Spatial Hazards and Disasters Database) \cite{sheldus} for different billion-dollar events in 2017, which indicates the losses in dollars across different counties. We choose to estimate the property damage and pose the task as a classification problem. The counties were classified into those with damages less than 10,000\$ (low severity), 10,000\$-100,000\$ (medium), and 100,000\$ and above (high). 

For DAG1, we used different tree-based and neural models to estimate property losses. For evaluation, we observe the best macro-F1 performance on a 5-fold cross-validation set. A 5-layer MLP was found to be the best-performing model. We use the best-performing model for all subsequent analyses (MLP 2-Layer). For DAG2 and DAG3, we seek to learn an SCM from the observational data, given that we have the structure defined. To implement SCMs, we trained a conditional probability model for every node (feature) based on its parent nodes. To generate output, we use all the models at every node, topologically sorted, to generate the classification.

\begin{figure*}[htbp!]
    \centering        \includegraphics[width=0.9\textwidth]{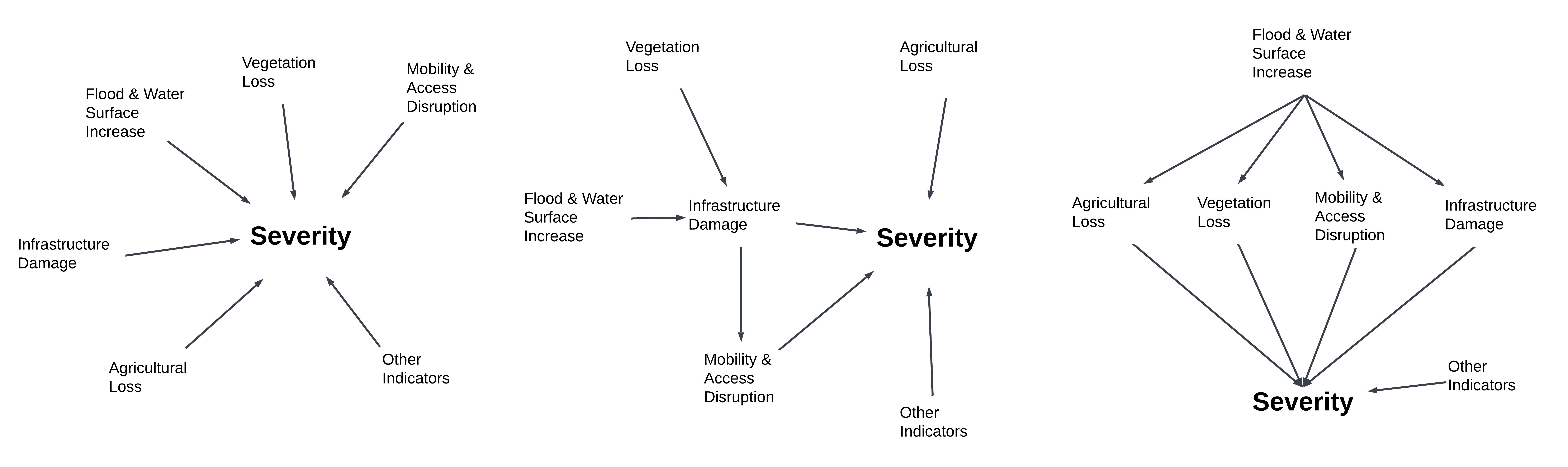} % replace with your filename
    \caption{Three different DAGs used to build interventional models.}
    \label{fig:dags}
\end{figure*}

%\vspace{-3 mm}

% \begin{table}[htbp]
% \centering
% \caption{Representative feature groupings for damage assessment. Here, t\_<source>\_<target> denotes area transitions between land covers during the disaster.}
% \begin{tabular}{p{0.27\columnwidth} p{0.65\columnwidth}}
% \toprule
% \textbf{Feature Group} & \textbf{Features} \\
% \midrule

% Vegetation Loss &
% t\_trees\_to\_water, news\_trees, reddit\_trees, t\_grass\_to\_bare, t\_shrubs\_to\_bare, ...\\
% \addlinespace

% Agricultural Loss &
% news\_agriculture, reddit\_agriculture, t\_crops\_to\_water,  t\_crops\_to\_bare, ... \\
% \addlinespace

% Infrastructural Damage &
% t\_built\_to\_water, t\_built\_to\_trees, reddit\_buildings, news\_roofs, reddit\_roofs, news\_power\_lines, reddit\_power\_lines, news\_infrastructure, ... \\
% \addlinespace

% Mobility \& Access &
% news\_vehicles, reddit\_vehicles, ... \\
% \addlinespace

% Flood \& Water &
% water\_surface\_area, t\_bare\_to\_water, ... \\
% \addlinespace

% Other Indicators &
% t\_water\_to\_bare, t\_bare\_to\_trees, ... \\
% \bottomrule
% \end{tabular}
% \label{tab:feature_groupings_minimal}
% \end{table}

%\vspace{-5mm}

\section{Overview of the tool}

Our tool is designed to support decision-makers by offering three core functionalities that combine interventional modelling, model interpretability, and recourse recommendations.

\subsection{Simulation through Interventions}

We provide an interactive interface where users can visualize county-level severity predictions of disaster severity. Additionally, the user has access to a dropdown menu to select various features that are preset to current values, allowing them to toggle the values of different features to generate a new assessment. From any of the three pre-defined DAGs, the users can select one of them to generate simulations for the chosen county. The users can then intervene on different factors (e.g., ``area change from built-up to water"). Therefore, the corresponding simulation would sever the effect of parent variables in generating the hypothetical assessment with the intervention. From a given SCM that we learn, we follow abduction, action, and prediction paradigm \cite{pearl2009causality} to generate counterfactuals. For the intervention step, we sever off relationships from all the parent features (features in the parent group) to the given feature.

\begin{figure*}[htbp]
    \centering    
    \includegraphics[height=8.5cm, width=0.8\textwidth, keepaspectratio=false]{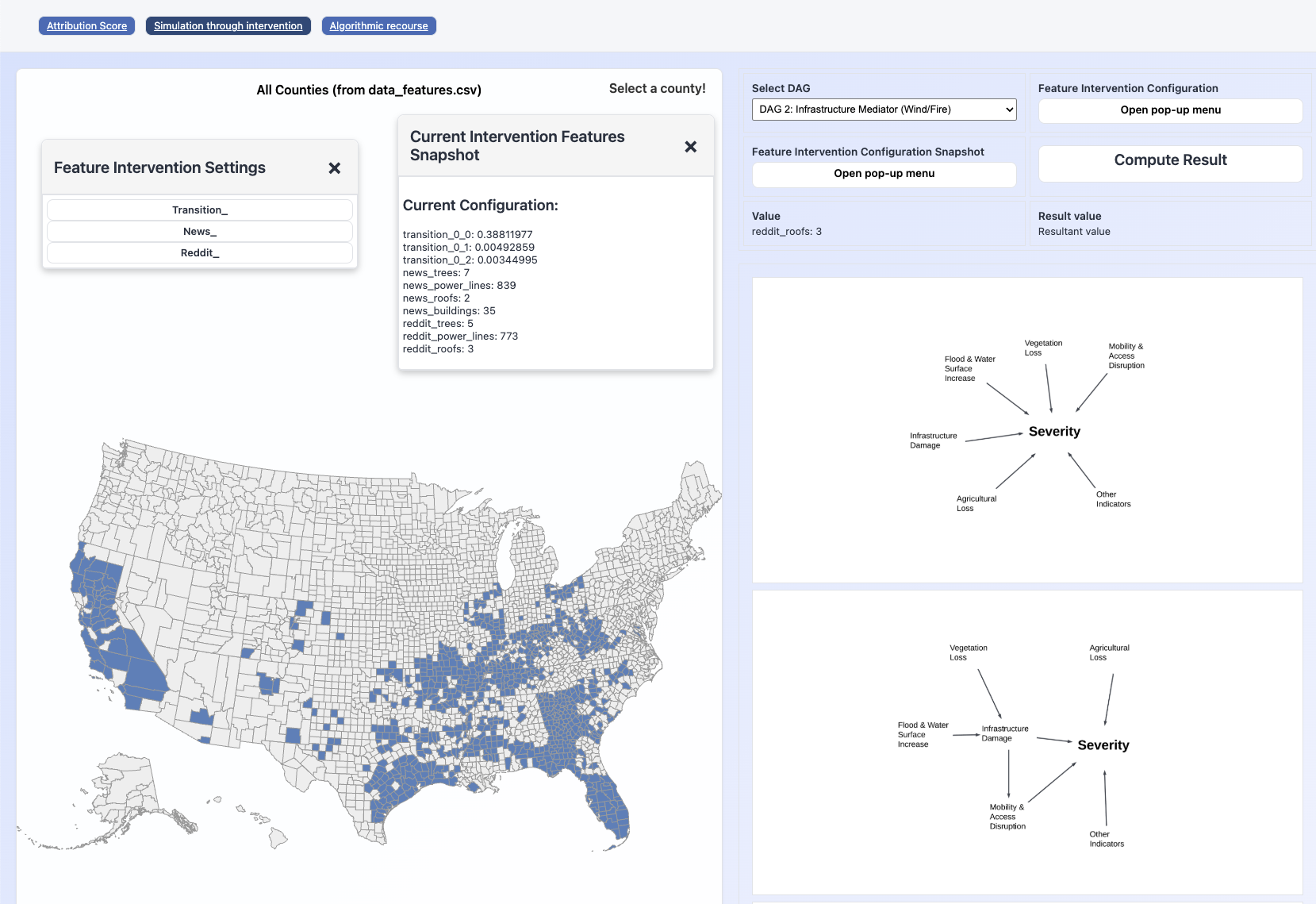}
    \caption{The above figure shows the simulation page, where we can intervene on different DAGs to generate assessments.}
    \label{fig:demo_view}
\end{figure*}

% \begin{figure*}[htbp]
%     \centering    
%     \includegraphics[width=0.7\textwidth]{figures/demo-app.png}
%     \caption{The above figure shows the simulation page, where we can intervene on different DAGs to generate assessments.}
%     \label{fig:demo_view}
% \end{figure*}

\begin{figure*}[htbp]
    \centering       \includegraphics[width=0.77\textwidth]{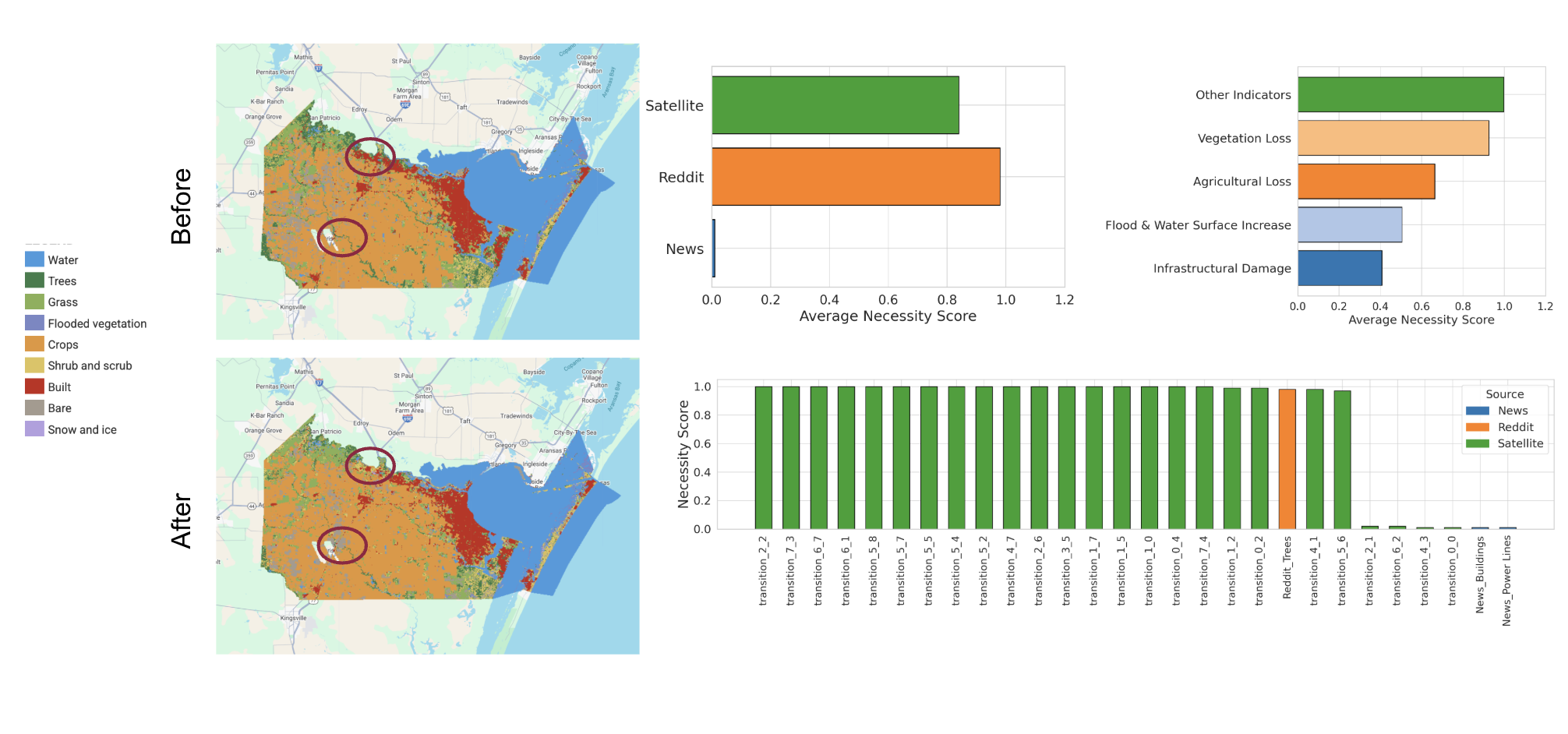} % replace with your filename
    \caption{Necessity score distributions for Hardin County, Texas, affected by Tornadoes in 2017. Red circles indicate a huge change in built-up area and bare land.}
    \label{fig:necessity_scores_yell}
\end{figure*}

\subsection{Attribution Scores of Different Factors}

% \begin{figure}[H]
%     \centering    
%     \includegraphics[width=1\columnwidth]{figures/necessity_scores_distribution_overall.pdf} % replace with your filename
%     \caption{Distribution of necessity attributions.}
%     \label{fig:necessity_importances}
% \end{figure}

\begin{figure}[H]
    \centering        \includegraphics[width=1\columnwidth]{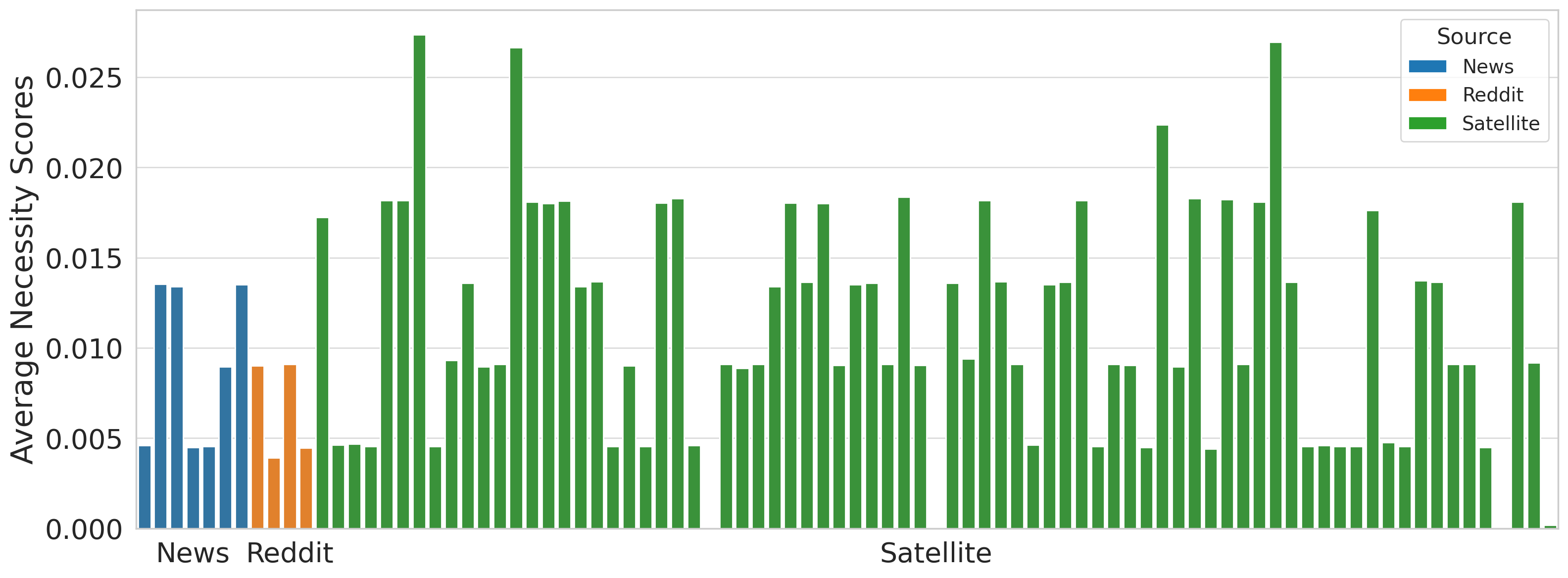} % replace with your filename
    \caption{Distribution of average necessity attribution scores across different features.}
    \label{fig:necessity_importances}
\end{figure}

To foster transparency and understanding, we provide attribution scores at three granularities: (a) \textit{source-level} (e.g., Reddit, News, Satellite), (b) \textit{individual feature-level}, and (c) \textit{group-level} (e.g., infrastructure damage, etc. When a user clicks on a specific county, we highlight the features most \textit{necessary} to that prediction—i.e., those that, if changed, could potentially alter the model’s output. This notion is derived from Actual Causality \cite{halpern2016actual}. Recently, there has been an adaptation of necessity scores for understanding counterfactual explanations\cite{mothilal2021towards}, which we follow.\\
\textbf{Necessity.} A feature (or subset of features) $\mathbf{x}_j = a$ is \emph{necessary} for a model output $y^* = f(\mathbf{x})$ if changing its value (while keeping other features fixed) leads to a different prediction. Formally, necessity ($\alpha$) of a given feature $x_j$ is quantified as:
\begin{equation}
\alpha = \mathbb{P}( f(\mathbf{x}) \ne f(\mathbf{x}_{-j}, x_j') \mid x_j = a,\ f(\mathbf{x}) = y^* ),
\end{equation}
where $\mathbf{x}_{-j}$ denotes all features except $x_j$, and $x_j'$ represents a counterfactual value. To implement the necessity scores, we generate counterfactual explanations and check the proportion of valid explanations (output flipped) generated by keeping the features fixed. Figure~\ref{fig:necessity_importances}, shows the distribution of necessity scores over different features.

In our application, upon selecting a given county, the user can see the 20 most necessary features in terms of necessity. The user has an option to further look at the attribution scores for different sources and different feature groups.

%\\
% \textbf{Sufficiency.}
% A feature (or subset of features) $\mathbf{x}_j = a$ is \emph{sufficient} for the output $y^* = f(\mathbf{x})$ if, when fixed, it ensures the same output regardless of the values of other features. This is captured by:
% \[
% \beta = \Pr\left( y = y^* \mid \mathbf{x}_j \leftarrow a \right)
% \]
% where $\mathbf{x}_j \leftarrow a$ denotes an intervention that sets $\mathbf{x}_j$ to $a$ in the model.
%Similarly, we calculate the sufficiency of each feature-i.e., if that feature is enough to arrive at the given prediction.

\subsection{Algorithmic Recourse for Decision Support}

% \begin{figure}[H]
%     \centering            \includegraphics[width=.7\columnwidth]{figures/recourse.pdf} % replace with your filename
%     \caption{Recourse through counterfactual explanation.}
%     %\vspace{-3 mm}
%     \label{fig:recourse}
% \end{figure}

Algorithmic recourse refers to the set of actions for the user to take to change the desired outcome \cite{karimi2021algorithmic}. Our tool enables users to generate recourse recommendations to change the current severity assessment, by suggesting modifications to a set of feature values. On the tool, users can select a desired outcome (e.g., low) against the current assessment (e.g., high). The output provides a lollipop chart indicating the set of suggested feature changes. To generate recourses, we use DiCE as a counterfactual explanation method. We add feasibility constraints (e.g., positivity of all features) and restrict the maximum number of features that can be changed.

%(see figure~\ref{fig:recourse})

% \subsection{Feature Importances - Necessity/Sufficiency}

\subsection{System Implementation}

The web application can be run locally by any user on their system. With a Flask backend, we use D3.js for visualizations. The webapp provides different pages for each of the three mentioned functionalities. In every page, we provide a map viewer that lets the user choose different counties for further analysis. In the first page (see Figure~\ref{fig:demo_view}), we let the user intervene by providing sliders against each feature that a user can use to change the current values and run a simulation to generate the new assessment. There is an additional page for visualizing attribution scores at the feature, feature-group, and source levels. For the feature-group importance and source-level importance, we display information for every feature-grouping or source, while we limit display to top 20 necessary features in the attribution view. In addition, we provide a third window, where the user for a given county of interest can set a desired state and then observe the course of action needed to change the outcome to the desired state. This would potentially inform first responders where they would need to intervene to change the current severity.

% \section{Case study}
% Our dataset comprises events from different disasters. In particular, our qualitative analysis indicates that Reddit as a data source might be more important. While the relevance for composite events does not give very interpretable importance, relevant features like water transitions for floods and tree/shrub-based transitions for wildfires turn out to be necessary.

%\vspace{-3 mm}

\section{Conclusion}

In this paper, we built a proof-of-concept tool to aid first responders during disaster response. Our tool helps users understand the importance of each feature in classifying the county into severity levels at feature, feature-group, and source levels. In addition, we provide the option of algorithmic recourse, suggesting suitable responses for an intervention. In our future work, we would want to extend this tool to gather real-time data streams.

%%
%% The acknowledgments section is defined using the "acks" environment
%% (and NOT an unnumbered section). This ensures the roper
%% identification of the section in the article metadata, and the
%% consistent spelling of the heading.
\begin{acks}
This material is based upon work supported by, or in
part by the U.S. Army Materiel
Command under Grant Award Number W911NF-24-2-0175.
\end{acks}

\section*{Generative AI Disclosure}
Portions of this paper were prepared with the assistance of generative AI tools (e.g., OpenAI's ChatGPT) for tasks such as coding, generating figures. The authors reviewed and verified all AI-assisted content to ensure its accuracy and integrity. All substantive ideas, experimental design, and results originate from the authors.

%%
%% The next two lines define the bibliography style to be used, and
%% the bibliography file.
\bibliographystyle{ACM-Reference-Format}
\balance
\bibliography{references}

%%% -*-BibTeX-*-
%%% Do NOT edit. File created by BibTeX with style
%%% ACM-Reference-Format-Journals [18-Jan-2012].

\begin{thebibliography}{9}

%%% ====================================================================
%%% NOTE TO THE USER: you can override these defaults by providing
%%% customized versions of any of these macros before the \bibliography
%%% command.  Each of them MUST provide its own final punctuation,
%%% except for \shownote{} and \showURL{}.  The latter two
%%% do not use final punctuation, in order to avoid confusing it with
%%% the Web address.
%%%
%%% To suppress output of a particular field, define its macro to expand
%%% to an empty string, or better, \unskip, like this:
%%%
%%% \newcommand{\showURL}[1]{\unskip}   % LaTeX syntax
%%%
%%% \def \showURL #1{\unskip}           % plain TeX syntax
%%%
%%% ====================================================================

\ifx \showCODEN    \undefined \def \showCODEN     #1{\unskip}     \fi
\ifx \showISBNx    \undefined \def \showISBNx     #1{\unskip}     \fi
\ifx \showISBNxiii \undefined \def \showISBNxiii  #1{\unskip}     \fi
\ifx \showISSN     \undefined \def \showISSN      #1{\unskip}     \fi
\ifx \showLCCN     \undefined \def \showLCCN      #1{\unskip}     \fi
\ifx \shownote     \undefined \def \shownote      #1{#1}          \fi
\ifx \showarticletitle \undefined \def \showarticletitle #1{#1}   \fi
\ifx \showURL      \undefined \def \showURL       {\relax}        \fi
% The following commands are used for tagged output and should be
% invisible to TeX
\providecommand\bibfield[2]{#2}
\providecommand\bibinfo[2]{#2}
\providecommand\natexlab[1]{#1}
\providecommand\showeprint[2][]{arXiv:#2}

\bibitem[Brown et~al\mbox{.}(2022)]%
        {brown2022dynamic}
\bibfield{author}{\bibinfo{person}{Christopher~F Brown}, \bibinfo{person}{Steven~P Brumby}, \bibinfo{person}{Brookie Guzder-Williams}, \bibinfo{person}{Tanya Birch}, \bibinfo{person}{Samantha~Brooks Hyde}, \bibinfo{person}{Joseph Mazzariello}, \bibinfo{person}{Wanda Czerwinski}, \bibinfo{person}{Valerie~J Pasquarella}, \bibinfo{person}{Robert Haertel}, \bibinfo{person}{Simon Ilyushchenko}, {et~al\mbox{.}}} \bibinfo{year}{2022}\natexlab{}.
\newblock \showarticletitle{Dynamic World, Near real-time global 10 m land use land cover mapping}.
\newblock \bibinfo{journal}{\emph{Scientific Data}} \bibinfo{volume}{9}, \bibinfo{number}{1} (\bibinfo{year}{2022}), \bibinfo{pages}{251}.
\newblock


\bibitem[{Federal Emergency Management Agency}(2020)]%
        {fema2020hazus}
\bibfield{author}{\bibinfo{person}{{Federal Emergency Management Agency}}.} \bibinfo{year}{2020}\natexlab{}.
\newblock \bibinfo{booktitle}{\emph{HAZUS-MH Technical Manual}}.
\newblock FEMA.
\newblock
\newblock
\shownote{Available at \url{https://www.fema.gov/flood-maps/products-tools/hazus}}.


\bibitem[Halpern(2016)]%
        {halpern2016actual}
\bibfield{author}{\bibinfo{person}{Joseph~Y. Halpern}.} \bibinfo{year}{2016}\natexlab{}.
\newblock \bibinfo{booktitle}{\emph{Actual Causality}}.
\newblock \bibinfo{publisher}{MIT Press}, \bibinfo{address}{Cambridge, MA}.
\newblock
\urldef\tempurl%
\url{https://mitpress.mit.edu/9780262035026/actual-causality/}
\showURL{%
\tempurl}


\bibitem[Hazards and Institute(2023)]%
        {sheldus}
\bibfield{author}{\bibinfo{person}{Hazards} {and} \bibinfo{person}{Vulnerability~Research Institute}.} \bibinfo{year}{2023}\natexlab{}.
\newblock \bibinfo{title}{SHELDUS: Spatial Hazard Events and Losses Database for the United States}.
\newblock \bibinfo{howpublished}{\url{https://cemhs.asu.edu/sheldus}}.
\newblock
\urldef\tempurl%
\url{https://cemhs.asu.edu/sheldus}
\showURL{%
\tempurl}
\newblock
\shownote{Center for Emergency Management and Homeland Security, Arizona State University}.


\bibitem[Karimi et~al\mbox{.}(2021)]%
        {karimi2021algorithmic}
\bibfield{author}{\bibinfo{person}{Amir-Hossein Karimi}, \bibinfo{person}{Bernhard Sch{\"o}lkopf}, {and} \bibinfo{person}{Isabel Valera}.} \bibinfo{year}{2021}\natexlab{}.
\newblock \showarticletitle{Algorithmic recourse: from counterfactual explanations to interventions}. In \bibinfo{booktitle}{\emph{Proceedings of the 2021 ACM conference on fairness, accountability, and transparency}}. \bibinfo{pages}{353--362}.
\newblock


\bibitem[Mothilal et~al\mbox{.}(2021)]%
        {mothilal2021towards}
\bibfield{author}{\bibinfo{person}{Ramaravind~Kommiya Mothilal}, \bibinfo{person}{Divyat Mahajan}, \bibinfo{person}{Chenhao Tan}, {and} \bibinfo{person}{Amit Sharma}.} \bibinfo{year}{2021}\natexlab{}.
\newblock \showarticletitle{Towards Unifying Feature Attribution and Counterfactual Explanations: Different Means to the Same End}. In \bibinfo{booktitle}{\emph{Proceedings of the 2021 AAAI/ACM Conference on AI, Ethics, and Society (AIES)}}. \bibinfo{publisher}{ACM}, \bibinfo{pages}{489--501}.
\newblock
\href{https://doi.org/10.1145/3461702.3462597}{doi:\nolinkurl{10.1145/3461702.3462597}}


\bibitem[{NASA Earth Science Applied Sciences Program}(2023)]%
        {nasaDisasters2023}
\bibfield{author}{\bibinfo{person}{{NASA Earth Science Applied Sciences Program}}.} \bibinfo{year}{2023}\natexlab{}.
\newblock \bibinfo{title}{NASA Disasters Program}.
\newblock \bibinfo{howpublished}{NASA Website}.
\newblock
\newblock
\shownote{Available at \url{https://disasters.nasa.gov}}.


\bibitem[Pearl(2009)]%
        {pearl2009causality}
\bibfield{author}{\bibinfo{person}{Judea Pearl}.} \bibinfo{year}{2009}\natexlab{}.
\newblock \bibinfo{booktitle}{\emph{Causality}}.
\newblock \bibinfo{publisher}{Cambridge university press}.
\newblock


\bibitem[Vishnubhatla(2025)]%
        {vishnubhatla2025impact}
\bibfield{author}{\bibinfo{person}{Saketh Vishnubhatla}.} \bibinfo{year}{2025}\natexlab{}.
\newblock \emph{\bibinfo{title}{Impact Assessment of Natural Disasters: A Computational Approach}}.
\newblock \bibinfo{thesistype}{Master's\ thesis}. \bibinfo{school}{Arizona State University}.
\newblock


\end{thebibliography}

%%
%% If your work has an appendix, this is the place to put it.
\appendix

\clearpage
\end{document}